\documentclass[conference]{IEEEtran}

\usepackage[utf8]{inputenc} 

\usepackage[noadjust]{cite}

\usepackage{graphicx}
\usepackage{subfig}
\usepackage[usenames]{color}

\usepackage[export]{adjustbox}

\usepackage[section]{placeins}
\usepackage{float}

\usepackage{subfiles}
\usepackage{blindtext}

\usepackage{enumitem}

\usepackage{algorithm}
\usepackage{algorithmicx}
\usepackage{algpseudocode}

\usepackage{amsmath}
\usepackage{amsfonts}
\usepackage{mathtools}

\usepackage{textcomp}
\usepackage{multirow}

\usepackage[capitalise]{cleveref}

\ifCLASSINFOpdf

\else

\fi

\begin{document}

\title{Analysis of Genetic Algorithm on \\ Bearings-Only Target Motion Analysis}

\author{
\IEEEauthorblockN{Erdem Köse}
\IEEEauthorblockA{Department of Electronics Engineering, Gebze Technical University\\
Gebze-Kocaeli, TR-41400, Turkey\\
Email: ekose@gtu.edu.tr}
}


\maketitle

\begin{abstract}
Target motion analysis using only bearing angles is an important study for tracking targets in water. Several methods including Kalman-like filters and evolutionary strategies are used to get a good predictor. Kalman-like filters couldn't get the expected results thus evolutionary strategies have been using in this area for a long time. Target Motion Analysis with Genetic Algorithm is the most successful method for Bearings-Only Target Motion Analysis and we investigated it. We found that Covariance Matrix Adaptation Evolutionary Strategies does the similar work with Target Motion Analysis with Genetic Algorithm and tried it; but it has statistical feedback mechanism and converges faster than other methods. In this study, we compared and criticize the methods.
\end{abstract}

\IEEEpeerreviewmaketitle

\section{Introduction} \label{sec:introduction}

	Bearings-Only Target Motion Analysis (BO-TMA) \cite{aytun_bearing-only_2019} is an analysis method for submarines using sonar bearings received from the hydrophone array. In this study we have only bearings information to track the target; this makes finding an optimal method for the study very hard.
	
	Kalman filters \cite{kronhamn_bearings-only_1998} are very popular in Target Motion Analysis (TMA), but due to initial value and linearization problems, evolutionary strategies \cite{beyer_evolution_2002} are more feasible than Kalman like optimum filters. Genetic Algorithms \cite{beyer_evolution_2002, ince_evolutionary_2009},  Big Bang-Big Crunch Algorithm (BB-BC) \cite{erol_new_2006, genc_bearing-only_2008,tokta_target_2017} and Covariance Matrix Adaptation Evolutionary Strategies (CMA-ES) \cite{hansen_reducing_2003,hansen_cma_2016,sonmez_new_2017} are the evolutionary strategies which have been used for BO-TMA.
	
	Cost function is simply the Euclidean distance between observed bearing angle and calculated bearing angle \cite{ince_evolutionary_2009}. This cost function has complex calculations when getting calculated bearing angle. Because of that a new cost function \cite{genc_new_2010} developed by formulating problem as to fit target courses such that the bearing lines divide the course candidates to equal distances and assuming bearings were measured at same constant time intervals. Equidistant line distance difference cost function is faster but not immune to noise.

	In \cref{sec:tma} we'll introduce the BO-TMA and it's important functions, in \cref{sec:tmaga} we'll introduce Target Motion Analysis with Genetic Algorithm (TMAGA) and investigate it, in \cref{sec:cmaes} we'll introduce CMA-ES and it's algorithm, in \cref{sec:trials} using TMAGA trial dataset we'll show the performance of CMA-ES, then in \cref{sec:conclusion} we'll summarize the paper.
	
\section{Bearings-Only Target Motion Analysis (BO-TMA)} \label{sec:tma}
\begin{figure}[H]
		\centering
		\includegraphics[width=0.6\linewidth]{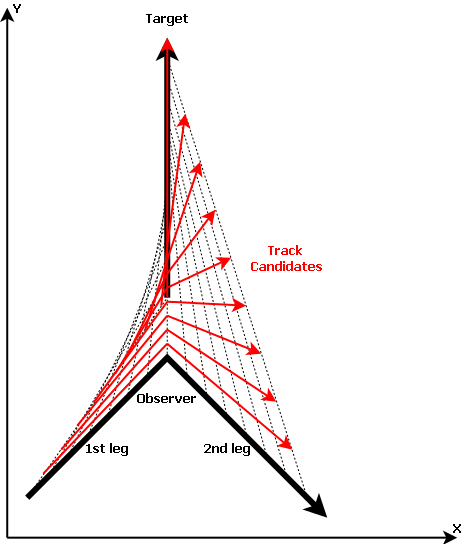}
		\caption{Possible track candidates for TMA.}
		\label{img:tma_scheme}
	\end{figure}

	In this section, we will introduce BO-TMA and essential equations to use evolutionary strategies for tracking target. As we mentioned before, BO-TMA is an analysis method for submarines using sonar bearings received from the hydrophone array. In \cref{img:tma_scheme} we see actual target with dashes, observer with straight bold line. Observer must make a leg to get only one possible linear track candidate for target's motion, otherwise there will be infinite possible linear track candidates. 
	
	\subsection{Position Equations}
		If we know our initial position $X_{obs}(0)$ and $Y_{obs}(0)$, target's initial bearing $B_{obsd}(0)$, initial range from observer $R_{tar}(0)$; then we can extract target's initial position $X_{tar}(0)$ and $Y_{tar}(0)$ from \cref{equ:target_motion_1}.
		
		\begin{equation}
			\left( 
			\begin{matrix}
				X_{tar}(0)\\
				Y_{tar}(0)
			\end{matrix}
			\right)
			=
			\left( 
			\begin{matrix}
				X_{obs}(0)\\
				Y_{obs}(0)
			\end{matrix}
			\right)
			+
			\left( 
			\begin{matrix}
				sin(B_{obsd}(0))\\
				cos(B_{obsd}(0))
			\end{matrix}
			\right)
			*
			R_{tar}(0)		
			\label{equ:target_motion_1}
		\end{equation}
		
		If we know our target's course angle $C_{tar}$ and velocity $S_{tar}$; then we can extract target's following positions  $X_{tar}(t+\Delta t)$ and $Y_{tar}(t+\Delta t)$, which is changing in $\Delta t$ equal time, from \cref{equ:target_motion_2}.
		
		\begin{equation}
			\left( 
			\begin{matrix}
				X_{tar}(t+\Delta t)\\
				Y_{tar}(t+\Delta t)
			\end{matrix}
			\right)
			=
			\left( 
			\begin{matrix}
				X_{tar}(t)\\
				Y_{tar}(t)
			\end{matrix}
			\right)
			+
			S_{tar}
			*
			\left( 
			\begin{matrix}
				sin(C_{tar})\\
				cos(C_{tar})
			\end{matrix}
			\right)
			*
			\Delta t		
			\label{equ:target_motion_2}
		\end{equation}

	\subsection{Cost Function}
		When we get the positions of target $X_{target_{EA}}(t)$ and $Y_{target_{EA}}(t)$ using evolutionary strategies, we can get estimated bearing angles using \cref{equ:bearing_calc}.
	
		\begin{equation}
			B_{estd}(t)
			=
			atan
			\left(
			\frac{X_{target_{EA}}(t)-X_{obs}(t)}
			{Y_{target_{EA}}(t)-Y_{obs}(t)}
			\label{equ:bearing_calc}
			\right)
		\end{equation}
		
		\begin{equation}
			Cost
			=
			\sqrt[2]{
				\sum_{i=0}^{N-1}{
					(B_{estd}(i)-B_{obsd}(i))^2
				}
			}
			\label{equ:cost_calc}
		\end{equation}
		
		The cost function for target motion analysis depends on Euclidean distance between estimated bearing angles $B_{estd}(t)$ and observed ones $B_{obsd}(t)$ is given by \cref{equ:cost_calc}. This cost function has more calculation complexity but more feasible than the equidistant line distance difference cost function \cite{genc_new_2010}.

\section{Target Motion Analysis with Genetic Algorithm (TMAGA)} \label{sec:tmaga}
	We may know all parameters except target's initial range from observer $R_{tar}(0)$, course of target $C_{tar}$ and velocity of target $S_{tar}$. Evolutionary strategies and genetic algorithms have been helping us for a long time to find these parameters in a short time and using a cost function. Exhaustive search can take so much time for large search areas; but evolutionary strategies gets help from statistics and cost functions to find true path to get true solution.

	TMAGA \cite{ince_evolutionary_2009} uses genetic algorithm to track target and has holding the best results in TMA studies. TMAGA has three principal structure; space narrowing and tuning chromosome lengths, N generation genetic algorithm and M times Monte Carlo runs. In each Monte Carlo run, 200 generation genetic algorithm runs for narrowing space then chromosomes are tuned using space limits; 500 generation genetic algorithm runs again to find best candidate. After 20 Monte Carlo runs system has 20 offspring to solve the problem. Then we take fitness weighted averages of these 20 offspring to get one solution. 
	
	In sections above, we will explain TMAGA in details and discuss the possible problems.

	\subsection{Target Motion Representation – Chromosome Structure}
		
		A solution individual is a chromosome bit array represents target motion. Solution individual's structure is given in \cref{equ:tmaga:chromosome}. $X_0$ and $Y_0$ are initial positions of target on $X$ and $Y$ axes respectively. $C$ is the course angle and $S$ is the velocity of target.

		\begin{equation}
			(...X_{0}...|...Y_{0}...|...C...|...S...)_2
			\label{equ:tmaga:chromosome}
		\end{equation}
		
		If we set a parameter bit-length for one of the four parameters above, we must know maximum $Max$ and minimum $Min$ value that parameter can get at limits. Then we use \cref{equ:tmaga:bit_width} with the degree of precision $p$, and the number of bits $m$. IEEE Standard for Floating-Point Arithmetic \cite{zuras_ieee_2008} says that for single precision floating point p parameter must be $log_{10}(2^{24}) \cong 7.225$. If we set precision 7 and $Max=20000$, $Min=0$ meters for $X_0$ and $Y_0$ parameters of chromosome, due to \cref{equ:tmaga:bit_width} our bit width will be 38. For $C$ and $S$ bit-widths are 32 and 28 due to limits [360 0] degrees and [25 0] meters/seconds respectively. Total 136 bits were used; but if we decide to use single precision floating point, we would use only 128 bits for representation.
		
		\begin{equation}
			2^{m-1} < (Max-Min) 10^p <2^m - 1
			\label{equ:tmaga:bit_width}
		\end{equation}
		
		Instead of representing position by known bearing $B_{obsd}$ and range $R$; the authors chose to use two parameters $X_0$ and $Y_0$. These values are correlated by the corresponding bearing. Then using two parameter; creates correlation problems, increases search dimension so enlarges search space causing more time to obtain correct solution. So ideal chromosome must look like \cref{equ:tmaga:chromosome_ideal}, this form uses 96 bits single precision floating point numbers.
		
		\begin{equation}
			(R_0|C|S)_{32bits~FP~Real~Value}
			\label{equ:tmaga:chromosome_ideal}
		\end{equation}
		
		In summary, genetic algorithm is a memory exhaustive and slow method to use in 2009, real valued evolutionary strategies uses real values instead of bit strings, which is more suitable for high level computing, and uses less memory.
	
	\subsection{Narrowing the Parameter Space}

		Due to TMAGA, search space between estimation and limit values could be reduced 20$\%$ using 200 generations with 50 population size and 20 runs. If we think that $\sigma=0.3$ in case of degrees, which is a quite high noise, generation and population size are quite low to converge to the ideal solution. Even if TMAGA says the parameter space's being narrowed into an infeasible area is a very unlikely case, it seems like very likely case, because genetic algorithm does not care the numerical value of the individual chromosome and does not use a feedback mechanism to narrow the area except the cost function. Because of this genetic algorithm results are expected to converge a different point in every different 20 runs. 
		
		If route is perpendicular to observer, noise becomes more destructive because determining the breaking point becomes more critical and it changes whole track even with a minimal change. In this case it is impossible to narrow the area to get correct result, because even with clean bearing it is hard to track target.
		
	\subsection{Cost Function}
		TMAGA uses a non-metric cost function called Total Deviation and shown in \cref{equ:tmaga:cost_function} where ${\Theta_G}_i$ and ${\Theta_M}_i$ are ground truth and calculated bearing angles of target at $i^{th}$ sample point. So this could cause divergence instead of convergence.
		
		\begin{equation}
			\text{Total Deviation}=\sum_{i=1}^{N}{\sqrt[2]{{\Theta_M^2}_i - {\Theta_G^2}_i}}
			\label{equ:tmaga:cost_function}
		\end{equation}
		
	 The ideal cost function is must be like in \cref{equ:tmaga:cost_function_ideal} to obtain Euclidean distance properties and become a metric distance measure.
	 
		\begin{equation}
			\text{Total Deviation}=\sqrt[2]{\sum_{i=1}^{N}{({\Theta_M}_i - {\Theta_G}_i)^2}}
			\label{equ:tmaga:cost_function_ideal}
		\end{equation}
		
	\subsection{Selection, Mutation and Crossover Operators}
		Genetic algorithm's mutation and crossover operators are somewhat weaker than real valued evolutionary strategy's ones. Because bitwise crossover and mutation options are limited and binary numbers' digit value is not meaningful by itself to these operators.
		
		TMAGA uses the simplest selection method; individuals are weighted by their fitnesses obtained from cost function, then randomly picked from population using these weights. The highest weighted individual has the highest pick chance, but the lowest weighted one has being picked chance too. But this straight method is weak; because the highest two individual's offspring won't create a good individual every time, sometimes two bad individual could create the best individual; so stochastic universal sampling and elitism could help converge faster than used method.
		
		TMAGA uses two piece crossover and randomly picked bitwise mutation methods. These methods are old fashioned for 2009 and their performances are not satisfying anymore. In real value measure non-uniform mutation and n-point real valued weighted averaging crossover methods could give much more better performance and easy to implement.
		
	\subsection{Function Evaluations}
\begin{figure}[H]
			\centering
			\includegraphics[width=\linewidth]{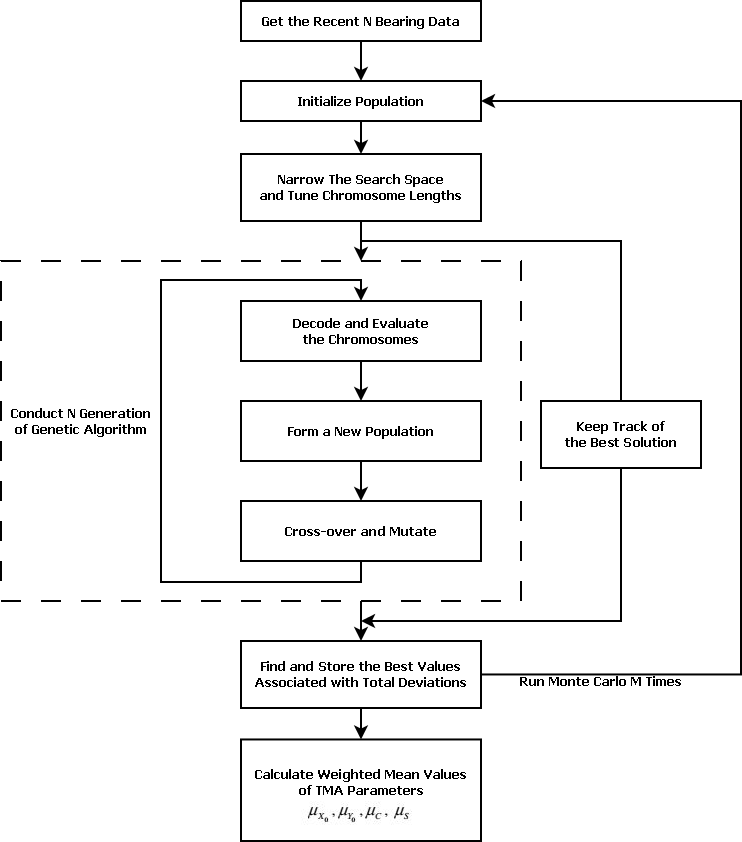}
			\caption{Main flow for TMAGA.}
			\label{img:tmaga_scheme}
		\end{figure}

		TMAGA's main flow include two layered structure. Outer layer creates population, narrows search space and keeps track of best solution. Inner layer controls genetic algorithm operations. The flow can be seen in \cref{img:tmaga_scheme}
		
		Outer and inner layers have Monte Carlo runs separately; so Monte Carlo run increases exponentially. TMAGA uses 20 inner, 20 outer Monte Carlo runs. Inner Monte Carlo is used for narrowing space and genetic algorithm run, outer one is used for statistical analysis.
		
		At the end of process, outer layer calculates weighted average of the best results from Monte Carlo runs and labels it as the best result.
		
		TMAGA uses 50 population size, 200 generations for space narrowing, 500 generations for genetic algorithm, 20 statistical, 20 in-system Monte Carlo counts. If we calculate function evaluation count with these informations; $50*700*20=700000$ function evaluation will be done in a single Monte Carlo of TMAGA, if 20 statistical Monte Carlo runs completed it will be $14000000$. 14 million function evaluation is too much for $limits(X_0)=[20000,0]$ meters, $limits(Y_0)=[20000,0]$ meters, $limits(C)=[360,0]$ degrees and $limits(S)=[25,0]$ meters/seconds.
		
		If we chose $(R_0|C|S)$ form \cref{equ:tmaga:chromosome_ideal} and $limits(R)=[28000,0]$ meters, $limits(C)=[360,0]$ degrees and $limits(S)=[25,0]$ meters/seconds; then search the space with 100 unit precision for $R$, 0.5 unit precision for $C$ and 0.1 unit precision for $S$ using brute force; $280*720*250=50400000$ function evaluation needed. Results for brute force searching is given in \cref{tab:tmaga:bruteforceresults} with ground truth values in \cref{tab:tmaga:groundtruth}. Even with $50400000$ function evaluations, we cannot get correct results.

		\begin{table}[H]
			\centering
			\caption{Trials' ground truth values.}
			\label{tab:tmaga:groundtruth}
				\begin{tabular}{|c|c|c|c|c|}
					\hline 
					\textbf{Trial} & \textbf{R(m)} & \textbf{C(\textdegree)} & \textbf{S(m/s)}& \textbf{$\sigma_{noise}$(\textdegree)}\\
					\hline 
					\textbf{1} & 7994 & 90 & 10 & 0.3 \\
					\hline 
					\textbf{2} & 7071 & 90 & 6 & 0.3 \\
					\hline 
					\textbf{3} & 4006 & 90 & 6 & 0.3 \\
					\hline 
					\textbf{4} & 8000 & 175 & 10 & 0.3 \\
					\hline 
					\textbf{5} & 8000 & 150 & 10 & 0.3 \\
					\hline 
					\textbf{6} & 8000 & 120 & 10 & 0.3 \\
					\hline 
					\textbf{7} & 4006 & 90 & 10 & 0 \\
					\hline 
					\textbf{8} & 4006 & 90 & 10 & 0.5 \\
					\hline 
					\textbf{9} & 4006 & 90 & 10 & 1 \\
					\hline 
					\textbf{10} & 4006 & 90 & 10 & 2 \\
					\hline 
					\textbf{11} & 4006 & 90 & 10 & 5 \\
					\hline 
					\textbf{12} & 4006 & 90 & 10 & 10 \\
					\hline 
				\end{tabular}
		\end{table}
			
		\begin{table}
			\centering
			\caption{Brute force search results.}
			\label{tab:tmaga:bruteforceresults}
			\begin{tabular}{|c|c|c|c||c|c|c|c|} 
				\hline
				\textbf{Trial}  & \textbf{R(m)}  & \textbf{C(\textdegree)}  & \textbf{S(m/s)}  & \textbf{Trial}~ ~ & \textbf{R(m)}  & \textbf{C(\textdegree)}  & \textbf{S(m/s)}   \\ 
				\hline
				\textbf{1}     & 7940           & 93            & 10               & \textbf{7}        & 4010           & 90            & 10                \\ 
				\hline
				\textbf{2}     & 6490           & 60            & 6                & \textbf{8}        & 8820           & 106           & 25                \\ 
				\hline
				\textbf{3}     & 4110           & 84            & 6                & \textbf{9}        & 5500           & 96            & 14                \\ 
				\hline
				\textbf{4}     & 4440           & 177           & 20               & \textbf{10}       & 6920           & 119           & 24                \\ 
				\hline
				\textbf{5}     & 7810           & 105           & 4                & \textbf{11}       & 3020           & 137           & 25                \\ 
				\hline
				\textbf{6}     & 10010          & 125           & 13               & \textbf{12}       & 11570          & 19            & 25                \\
				\hline
			\end{tabular}
		\end{table}
		
		From \cref{tab:tmaga:bruteforceresults} we can see that even brute force cannot find a clear solution for that high noise levels. Thus we can say noise is disrupting useful information in bearing data.
		
	\subsection{Statistical Results}
		We said that TMAGA has not guaranteed results before; because in the paper \cite{ince_evolutionary_2009} there is no information about variance of Monte Carlo runs and statistics are only given for means or the best results. Thus we cannot determine the real performance of the TMAGA. Higher performances on higher error standart deviations like $\sigma_{noise}=0.5$ and $\sigma_{noise}=1.0$ in degree measure are suspicious. Additionally, perpendicular movement of target to observer makes tracking difficult; but TMAGA says it is tracking target very accurate with these error variances. Even we look at \cref{tab:tmaga:bruteforceresults}, brute force couldn't get the precise results.
		
		If we look at Table 3 which is given by \cite{ince_evolutionary_2009}, we could see Monte Carlo is done two times and mean of the second Monte Carlo is given as the average result. There is no information about variance or standard deviation; so this type of reporting is not acceptable. Solutions are obtained by weighted averages of the first Monte Carlo's results. There is no statistical relation between best solution and fitness in extra noisy data condition. 

\section{Covariance Matrix Adaptation Evolutionary Strategies (CMA-ES)} \label{sec:cmaes}
	In this section, we will explain how the Covariance Adaptation Evolutionary Strategies \cite{hansen_cma_2016} works and model a new TMA system by CMA-ES.
	
	\subsection{CMA-ES}
	
	CMA-ES is a kind of evolutionary strategy that uses statistical feedback and covariance matrix for capturing search space shape. CMA-ES uses cost weighted recombination instead of crossover; random number generation using Gaussian distribution instead of mutation.
		
	$m_i$ is the favorite solution at the start of each generation and could be obtained using \cref{equ:cma:favorite_solution}. $w_{i:\lambda}$ is the weight obtained from the cost of the parent $i$ for the $\lambda^{th}$ offspring cluster, $x_i$ is the $i^{th}$ parent.
	
	\begin{equation}
		\begin{split}
			w_{1:\lambda}>=w_{2:\lambda}>=...>=w_{\mu:\lambda} &\Rightarrow \\
			\sum_{i=1}^{\mu}{w_{i:\lambda}}=1 &\Rightarrow m_i=\sum_{i=1}^{\mu}{x_i w_{i:\lambda}}
		\end{split}
		\label{equ:cma:favorite_solution}
	\end{equation}
	
	Covariance matrix $C$ determines the shape of the random number generation and step size $\sigma$ determines the width of probability density function. Initially we use $\sigma$ predefined around like a number 0.3 and $C$ unit diagonal matrix. We learn them each generation via the methods described by Hansen\cite{hansen_cma_2016}. Then we use \cref{equ:cma:new_offspring} to get new population individuals $x_i$ around the $m_i$ favorite individual.
	
	\begin{equation}
		x_i=m_i + \sigma_{CMA} N_i(0,C)
		\label{equ:cma:new_offspring}
	\end{equation}
	
\begin{figure}[H]
		\centering
		\includegraphics[width=1\linewidth]{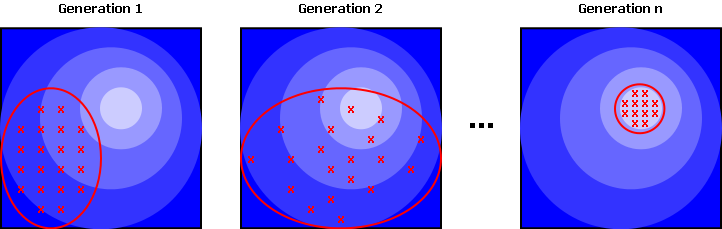}
		\caption{Concept work of the CMA-ES.}
		\label{img:cma:concept}
	\end{figure}

	CMA-ES narrows the search space using statistical information learned by updating $C$ and $\sigma$. This narrowing helps CMA-ES converge with less calculations than classical evolutionary strategies. \cref{img:cma:concept} shows how CMA works and converges to the result by the help of $C$ and $\sigma$.
	
\section{Trials} \label{sec:trials}
	We're using CMA \cite{hansen_cma_2016} with cost function given by \cref{equ:cost_calc} to solve the target motion analysis problem. CMA is a powerful method as it converges quickly to the solution without stucking to local minima. Our parameters for CMA is given in \cref{tab:cma:parameters}

	\begin{table}[H]
		\centering
		\caption{CMA parameters.}
		\label{tab:cma:parameters}
		\begin{tabular}{|c|c|}
			\hline 
			\textbf{Parameter} & \textbf{Value} \\ 
			\hline 
			\textbf{No of Parameters} & 3\\
			\hline 
			\textbf{Parent Size} & 100\\
			\hline 
			\textbf{Offspring Size} & 100\\
			\hline 
			\textbf{No of Generations} & 50000\\
			\hline 
			\textbf{Function Evaluation Budget} & 50000\\
			\hline 
			\textbf{Range Limits} & [0, 28000] meters\\
			\hline 
			\textbf{Course Angle Limits} & [0, 360] degrees\\
			\hline 
			\textbf{Speed Limits} & [0, 25] meters/second\\
			\hline 
			\textbf{No of Monte Carlo Runs} & 1000\\ 
			\hline 
		\end{tabular}
	\end{table}

	\begin{table}
		\centering
		\caption{CMA-ES search results.}
		\label{tab:cma:normalresults}
		\begin{tabular}{|c|c|c|c|c|} 
			\hline
			\multicolumn{2}{|c|}{\textbf{Trial} }              & \textbf{R(m)}  & \textbf{C(\textdegree)}  & \multicolumn{1}{c|}{\textbf{S(m/s)} }  \\ 
			\hline
			\multirow{3}{*}{\textbf{1} }  & \textbf{$\mu$}     & 9732.594       & 89.608        & 13.075                                 \\
			& \textbf{$\sigma$}  & 4619.759       & 19.677        & 6.662                                  \\
			& \textbf{$|Dev.|$}  & 1738.371       & 0.391         & 3.075                                  \\ 
			\hline
			\multirow{3}{*}{\textbf{2} }  & \textbf{$\mu$}     & 8503.276       & 88.662        & 9.710                                  \\
			& \textbf{$\sigma$}  & 4105.756       & 36.325        & 6.609                                  \\
			& \textbf{$|Dev.|$}  & 1432.208       & 1.337         & 3.710                                  \\ 
			\hline
			\multirow{3}{*}{\textbf{3} }  & \textbf{$\mu$}     & 4471.893       & 90.753        & 7.272                                  \\
			& \textbf{$\sigma$}  & 1648.329       & 18.434        & 3.632                                  \\
			& \textbf{$|Dev.|$}  & 465.098        & 0.753         & 1.272                                  \\ 
			\hline
			\multirow{3}{*}{\textbf{4} }  & \textbf{$\mu$}     & 9504.828       & 126.924       & 16.7323                                \\
			& \textbf{$\sigma$}  & 5064.392       & 83.3033       & 8.4811                                 \\
			& \textbf{$|Dev.|$}  & 1504.822       & 48.075        & 6.732                                  \\ 
			\hline
			\multirow{3}{*}{\textbf{5} }  & \textbf{$\mu$}     & 8273.251       & 123.913       & 13.074                                 \\
			& \textbf{$\sigma$}  & 2308.739       & 49.989        & 8.5072                                 \\
			& \textbf{$|Dev.|$}  & 273.244        & 26.086        & 3.074                                  \\ 
			\hline
			\multirow{3}{*}{\textbf{6} }  & \textbf{$\mu$}     & 9017.057       & 110.775       & 13.375                                 \\
			& \textbf{$\sigma$}  & 3679.193       & 28.754        & 8.153                                  \\
			& \textbf{$|Dev.|$}  & 1017.051       & 9.224         & 3.375                                  \\ 
			\hline
			\multirow{3}{*}{\textbf{7} }  & \textbf{$\mu$}     & 4006.794       & 90            & 10                                     \\
			& \textbf{$\sigma$}  & 0              & 0             & 0                                      \\
			& \textbf{$|Dev.|$}  & 0              & 0             & 0                                      \\ 
			\hline
			\multirow{3}{*}{\textbf{8} }  & \textbf{$\mu$}     & 5119.677       & 90.475        & 13.389                                 \\
			& \textbf{$\sigma$}  & 2436.746       & 11.958        & 7.017                                  \\
			& \textbf{$|Dev.|$}  & 1112.883       & 0.475         & 3.389                                  \\ 
			\hline
			\multirow{3}{*}{\textbf{9} }  & \textbf{$\mu$}     & 5368.986       & 87.240        & 14.419                                 \\
			& \textbf{$\sigma$}  & 2994.409       & 18.946        & 8.552                                  \\
			& \textbf{$|Dev.|$}  & 1362.192       & 2.759         & 4.419                                  \\ 
			\hline
			\multirow{3}{*}{\textbf{10} } & \textbf{$\mu$}     & 5360.204       & 83.377        & 14.792                                 \\
			& \textbf{$\sigma$}  & 3363.029       & 25.509        & 9.349                                  \\
			& \textbf{$|Dev.|$}  & 1353.410       & 6.622         & 4.792                                  \\ 
			\hline
			\multirow{3}{*}{\textbf{11} } & \textbf{$\mu$}     & 5517.113       & 77.005        & 15.927                                 \\
			& \textbf{$\sigma$}  & 4140.429       & 37.484        & 9.844                                  \\
			& \textbf{$|Dev.|$}  & 1510.319       & 12.994        & 5.927                                  \\ 
			\hline
			\multirow{3}{*}{\textbf{12} } & \textbf{$\mu$}     & 3877.869       & 90.121        & 16.445                                 \\
			& \textbf{$\sigma$}  & 5939.399       & 77.892        & 10.674                                 \\
			& \textbf{$|Dev.|$}  & 128.9249       & 0.121         & 6.445                                  \\
			\hline
		\end{tabular}
	\end{table}

	Results for CMA-ES searching is given in \cref{tab:cma:normalresults} with ground truth values in \cref{tab:tmaga:groundtruth}. $\mu$ represents Monte Carlo mean, $\sigma$ represents Monte Carlo standart deviation and $|Dev.|$ represents the absolute deviation between estimation and ground truth. 	As we can see from \cref{tab:cma:normalresults}, author's error standart deviation is too much for tracking and result variates in wide ranges. If we reduce error standart deviation by 0.1x, we can see that results are becoming much more reliable. We can say that maximum error standart deviation is $\sigma_{noise}=0.2$\textdegree.

	CMA-ES solves problem with $50000$ function evaluation which is $14$ times smaller than $700000$ function evaluation of brute force and results' standart deviation is smaller enough to trust.
	
\section{Conclusion} \label{sec:conclusion}
	As conclusion we can say that genetic algorithm is a non-modern weak method to solve target motion analysis. Even using two different position parameters enlarges search space and creates correlation problems. Distance measure is wrong; because it don't justify metric rules. Search space narrowing is useless because algorithm may converge to local minima with genetic algorithm, there is no feedback to avoid it. Statistical results are given in wrong way, and there is too much calculation to solve problem.

	CMA-ES narrows space via statistical feedback and converges quickly without being stuck to local minima. Error standart deviation is too large to converge, with experiments we can say 0.2 is the maximum value for error standart deviation. As result we can get solution with $14$ times smaller function evaluation count than brute force via CMA-ES. 
	
\bibliographystyle{IEEEtran}
\bibliography{bibtma}

\begin{thebibliography}{10}
\providecommand{\url}[1]{#1}
\csname url@samestyle\endcsname
\providecommand{\newblock}{\relax}
\providecommand{\bibinfo}[2]{#2}
\providecommand{\BIBentrySTDinterwordspacing}{\spaceskip=0pt\relax}
\providecommand{\BIBentryALTinterwordstretchfactor}{4}
\providecommand{\BIBentryALTinterwordspacing}{\spaceskip=\fontdimen2\font plus
\BIBentryALTinterwordstretchfactor\fontdimen3\font minus
  \fontdimen4\font\relax}
\providecommand{\BIBforeignlanguage}[2]{{%
\expandafter\ifx\csname l@#1\endcsname\relax
\typeout{** WARNING: IEEEtran.bst: No hyphenation pattern has been}%
\typeout{** loaded for the language `#1'. Using the pattern for}%
\typeout{** the default language instead.}%
\else
\language=\csname l@#1\endcsname
\fi
#2}}
\providecommand{\BIBdecl}{\relax}
\BIBdecl

\bibitem{aytun_bearing-only_2019}
A.~Aytun and S.~Bulkan, ``Bearing-{Only} {Target} {Motion} {Analysis},'' in
  \emph{Operations {Research} for {Military} {Organizations}}.\hskip 1em plus
  0.5em minus 0.4em\relax IGI Global, 2019, pp. 330--346.

\bibitem{kronhamn_bearings-only_1998}
T.~Kronhamn, ``Bearings-only target motion analysis based on a multihypothesis
  {Kalman} filter and adaptive ownship motion control,'' \emph{IEE
  Proceedings-Radar, Sonar and Navigation}, vol. 145, no.~4, pp. 247--252,
  1998.

\bibitem{beyer_evolution_2002}
H.-G. Beyer and H.-P. Schwefel, ``Evolution strategies{\textendash}{A}
  comprehensive introduction,'' \emph{Natural computing}, vol.~1, no.~1, pp.
  3--52, 2002.

\bibitem{ince_evolutionary_2009}
L.~Ince, B.~Sezen, E.~Saridogan, and H.~Ince, ``An evolutionary computing
  approach for the target motion analysis ({TMA}) problem for underwater
  tracks,'' \emph{Expert Systems with Applications}, vol.~36, no.~2, pp.
  3866--3879, 2009.

\bibitem{erol_new_2006}
O.~K. Erol and I.~Eksin, ``A new optimization method: big bang{\textendash}big
  crunch,'' \emph{Advances in Engineering Software}, vol.~37, no.~2, pp.
  106--111, 2006.

\bibitem{genc_bearing-only_2008}
H.~Genc and A.~Hocaoglu, ``Bearing-only target tracking based on big
  bang{\textendash}big crunch algorithm,'' in \emph{Computing in the {Global}
  {Information} {Technology}, 2008. {ICCGI}'08. {The} {Third} {International}
  {Multi}-{Conference} on}.\hskip 1em plus 0.5em minus 0.4em\relax IEEE, 2008,
  pp. 229--233.

\bibitem{tokta_target_2017}
A.~Tokta, A.~K. Hocao{\u g}lu, and H.~M. Gen{\c c}, ``Target motion analysis
  with fitness adaptive big bang-big crunch optimization algorithm,'' in
  \emph{Signal {Processing} and {Communications} {Applications} {Conference}
  ({SIU}), 2017 25th}.\hskip 1em plus 0.5em minus 0.4em\relax IEEE, 2017, pp.
  1--4.

\bibitem{hansen_reducing_2003}
N.~Hansen, S.~D. M{\"u}ller, and P.~Koumoutsakos, ``Reducing the time
  complexity of the derandomized evolution strategy with covariance matrix
  adaptation ({CMA}-{ES}),'' \emph{Evolutionary computation}, vol.~11, no.~1,
  pp. 1--18, 2003.

\bibitem{hansen_cma_2016}
N.~Hansen, ``The {CMA} evolution strategy: {A} tutorial,'' \emph{arXiv preprint
  arXiv:1604.00772}, 2016.

\bibitem{sonmez_new_2017}
H.~H. S{\"o}nmez, A.~K. Hocao{\u g}lu, and H.~M. Gen{\c c}, ``A new solution
  method for the passive target motion analysis problem using evolutionary
  strategies,'' in \emph{Signal {Processing} and {Communications}
  {Applications} {Conference} ({SIU}), 2017 25th}.\hskip 1em plus 0.5em minus
  0.4em\relax IEEE, 2017, pp. 1--4.

\bibitem{genc_new_2010}
H.~Gen{\c c}, ``A new solution approach for the bearing only target tracking
  problem,'' in \emph{Soft {Computing} {Applications} ({SOFA}), 2010 4th
  {International} {Workshop} on}.\hskip 1em plus 0.5em minus 0.4em\relax IEEE,
  2010, pp. 95--100.

\bibitem{zuras_ieee_2008}
D.~Zuras, M.~Cowlishaw, A.~Aiken, M.~Applegate, D.~Bailey, S.~Bass,
  D.~Bhandarkar, M.~Bhat, D.~Bindel, S.~Boldo, and {others}, ``{IEEE} standard
  for floating-point arithmetic,'' \emph{IEEE Std 754-2008}, pp. 1--70, 2008.

\end{thebibliography}

\end{document}